# Machine Learning Markets

**Amos Storkey**
School of Informatics, University of Edinburgh
a.storkey@ed.ac.uk

## Abstract

Prediction markets show considerable promise for developing flexible mechanisms for machine learning. Here, machine learning markets for multivariate systems are defined, and a utility-based framework is established for their analysis. This differs from the usual approach of defining static betting functions. It is shown that such markets can implement model combination methods used in machine learning, such as product of expert and mixture of expert approaches as equilibrium pricing models, by varying agent utility functions. They can also implement models composed of local potentials, and message passing methods. Prediction markets also allow for more flexible combinations, by combining multiple different utility functions. Conversely, the market mechanisms implement inference in the relevant probabilistic models. This means that market mechanism can be utilized for implementing parallelized model building and inference for probabilistic modelling.

## 1 Introduction

One intriguing feature of the history of machine learning, is that despite its ubiquitous methods, its immediate importance in a data rich world, and the desire for automation, the machine learning endeavour typically involves tackling each new problem through the individual crafting of a solution by experienced practitioners. The practitioners often use compositional structures to build machine learning models, but despite the large number of different models, the number of different compositional approaches is quite small. Almost all fall in to one or other of the following categories:

**Model averaging** e.g. Bayesian model averaging, boosting [9, 20].

**Mixtures** e.g. mixture of experts [13], mixture models, topic models [12, 5] and Dirichlet process mixtures.

**Products/Factors** e.g. Markov random fields, Conditional random fields, product of experts [11], Boltzmann machines, belief networks.

**Mixings** e.g. Independent Component Analysis based models and Gaussianization [21].

Although hybrid modelling has an extensive history, it is still the case that individual models are usually a composition of multiple *homogenous* elements rather than inhomogeneous ones. Despite this, the results of the Netflix challenge [3] suggest these individually designed results seem to be outperformed by combinations of differing methods, pooled using fairly simple pooling (e.g. model averaging) mechanisms.

Extending machine learning methods to more and more complicated scenarios will require increasing the flexibility of the modelling approaches. It may well be desirable to build models from inhomogeneous units as standard, and experiment with more flexible compositional methods. In this paper we suggest that *machine learning markets* play a role in this. If they are to contribute in this way then we must establish that they can extend machine learning methods. At the very least such markets must be able to implement current model structures and perform inference in those structures. A firm probabilistic interpretation is important, and any new approach should allow both the freedom for individual model building and the suitable combination of methods. In the paper we show that given a set of agents, the market equilibrium can implement a number of standard componential probabilistic model formalisms. Hence market dynamics provide a mechanism for probabilistic inference in those models.

Machine Learning Markets are prediction markets involving individual machine learning agents, each with a utility function and a probabilistic belief about the domain to be modelled. The goods in this market represent bets on outcomes of individual system states, and a no-arbitrage assumption means that the price of goods can be interpreted probabilistically. The approach of this prediction market goes beyond the simple single binary state predictors that are common in consideration of real prediction markets.

Rather we consider prediction markets on large joint spaces and multiple variables at one time.

We establish that some of the key model combination methods listed above can be directly implemented by groups of agents, and inference in those models is obtained by market dynamics. Different model combination methods are obtained via different utility functions for the individual agents. We also show that it is possible to consider agents with niche beliefs about a small subset of the variables in the system, and show that we can derive typical factor graph (i.e. product) representations from such systems.

As the agents are autonomous entities acting in a market, the methods outlined here are very amenable to parallelism. We show that by restricting the goods available in the market, the market dynamics can be represented as message-passing mechanisms between pricing and stock-holding, which are akin to the messages between variables and factors in message passing on factor graphs.

Most importantly the homogeneity of the agents which form the above models can be relaxed in favour of many different forms of agents with varying utility functions, without any change in the overall structure of the system. As a result a whole spectrum of different model combination procedures can be implemented here.

The focus of this paper is the examination of model combination methods for agents that have already learnt their beliefs. We leave the examination of learning in these markets to future work. However this examination of combination methods is also of moot philosophical interest. Subjective Bayesian methods for handling the updating of individual belief are well established. However the issue of rationally combining the posterior beliefs of *different* agents to form a consensus belief is a long-standing and unsolved issue in Bayesian philosophy [10]. Though we do not pretend that the approach described here is a solution to this, we do suggest that it is one way the problem could be considered, and indeed relates to the considerations given by other authors [19, 18, 14].

Finally we note that one further issue that is not discussed in any detail in this paper, but is nevertheless important. The flexibility of the market structure allows any agent to produce new derivative stocks which can be added to the market, and can be traded on like any other. This effectively allows for the generation of new features.

In the market economic analysis, we will take a neoclassical perspective and primarily utilize a competitive equilibrium assumption. This assumption is merely for illustration of the consequences of such equilibria, rather than suggesting that is precisely how such a market would operate. Because we will be utilizing concave utility functions, we know the market system will have a unique fixed point [1].

## 2 Previous Work

Here we summarize the main previous work in machine learning and prediction markets. The number of papers directly establishing market mechanisms for implementing existing probabilistic machine learning methods is small. One important paper is [15], where the authors do consider a prediction market to form model combination for machine learning. They only consider predictions regarding a single multinomial variable, and their agents do not have utilities, but are instead endowed with betting functions. These betting functions may not be derivable from any suitable utility, and indeed require that the amount bet is proportional to the total wealth, a constraint not seen by some of the utility functions used here. Even so this paper establishes that artificial prediction markets can provide useful mechanisms for combining classifiers. We believe the power of prediction markets can go well beyond this and can be a powerful tool for the overall machine learning endeavour. More recently other learning methods have been related to prediction markets [7]. However here the focus is on cost function based markets, where a global market maker is defined and the global cost function for the market is specified. The case we consider here is a more general market condition with no global market marker and independent agents, defined by their respective utility functions. Each agent follows a standard utility maximizing procedure: this is fully parallel. We also consider the case where goods can correspond to only a limited number of the (usually exponential number of) marginal outcomes, which has not been discussed hitherto.

The potential of prediction markets has long been talked about [2, 17, 24]. In [8] the authors compared a number of different mechanisms for expert aggregation including a simple prediction market approach. Different market designs have different features, and ensuring good prediction market design with sufficient fluidity [6] will be critical for efficiently reaching equilibrium. In [22] the authors examine the statistical properties of market agent models, whereas in [16] the authors consider prediction markets in the context of Bayesian learning.

## 3 Prediction Markets

A market provides a basic process for the exchange of goods between different agents. We define a basic market as follows.

**Definition 1 (Market)** *A market is a mechanism for the exchange of goods. The market itself is neutral with respect to the goods, or the trades. As such the market itself cannot acquire or owe goods, and hence is subject to the* market constraint *that the total number of goods sold is equal to the total number of goods bought. Any currency is simply another good in the context of a basic market. However we will assume that in this context there is an agreed cur-* rency: *all participants in the market are happy to use that currency for the purposes of trade.*

Within this definition, there are many potential forms and mechanisms for implementing a market. For the sake of simplicity will only consider charge-free markets in the context of this paper. We define a *position* in this market as the stock holding (i.e. number of each of the goods owned) of any individual. We assume that agents can hold a *short position* (i.e. debt or negative holding) in a stock. A positive holding in a stock is called a *long position*.

Suppose we have a market where one type of good being traded is a bet, and the other is a currency. In this context we can define a bet as a good that pays a fixed amount (taken to be 1 Grubnick[1] without loss of generality) dependent on a particular outcome of a future occurrence, and pays nothing otherwise. Markets consisting of trades of this form of good are called *prediction markets*[2].

**Definition 2 (Prediction Market)** *A prediction market (for the purposes of this paper) is a market with an agreed currency and where the remaining goods are bets with a fixed return on a particular outcome of a future occurrence (and a zero return otherwise). Individuals may choose to create those goods for sale, i.e. produce a bet and sell it at a price. This is equivalent to a short position in that bet.*

We will also make the assumption that the agents in the market view the currency as a risk-free asset, in that they are happy to define utility functions in terms of that currency.

## 4 Definitions

We start with a basic definition of the terms we will be using, followed by examples of how these will actually be used in practice.

Suppose we have a sample space $\Omega$ of all possible outcomes of the set of relevant (future) occurrences. The elements of $\Omega$ are called *events*, and one and only one of those events will be the actual *outcome*.

Suppose we also have a $\sigma$-field $\mathfrak{F}$ on $\Omega$. We enumerate a set of *market goods* by $k = 1, 2, \ldots N_G$, each associated with a set $m_k \in \mathfrak{F}$ to be bets that pay out 1 Grubnick if the outcome is in $m_k$.

We enumerate a set of agents $i = 1, 2, \ldots N_A$. Each agent can buy or sell any of the market goods. Hence each agent has a *position vector* (or stock holding) $\mathbf{s}_i$ in all the goods available. $s_{ik}$ is the total number of items agent $i$ has of good $k$. $s_{ik} < 0$ indicates a short position in that good. Note that is this paper $s_{ik}$ is not the total amount invested in item $k$: that depends also on the costs of the good.

Each agent also has an associated utility function $U_i(W)$

---

[1] The currency of Elbonia is, naturally, respected worldwide.

[2] More specifically this is sometimes called a winner-takes-all market: see e.g. [24, p2].

defined in terms of the currency, denoting the utility to the agent of a wealth of $W$ Grubnicks. Each agent will also have a belief, that is a probability measure $P_i$ defined on $(\Omega, \mathfrak{F})$. We can also consider agents who have beliefs defined on subspaces $(\Omega_i, \mathfrak{F}_i)$ of the probability space $(\Omega, \mathfrak{F})$. We call these *local beliefs*, and this will be appropriate for example where we consider distributions of many random variables, and these sub-fields are the $\sigma$-fields induced by certain subsets of those random variables.

In practice we will work with random variables, and hence the underlying $\sigma$-fields will be implicit. We will consider the cases of a single multiclass random variable, and a discrete multivariate random variable. We will define specific market goods in each instance.

Although many different utility functions are possible, three will be particularly important in this paper. These are now given.

### 4.1 Various Utility Functions

#### 4.1.1 Linear debt-free utility

The first utility function we will consider is
$$U^S(x) = x \text{ if } x > 0 \text{ and } -\infty \text{ otherwise.} \quad (1)$$
(where S denotes straight). This utility function prevents an agent from going into debt but otherwise is linear. This utility is not strictly concave (concave utility functions result in equilibrium solutions).

#### 4.1.2 Logarithmic utility

The second utility is concave, and takes a logarithmic form. This too does not allow debt, but has decreasing utility gains for increasing wealth.
$$U^L(x) = \log x \quad (2)$$
where $L$ is for logarithm.

#### 4.1.3 Exponential decaying negative utility

The third utility function that we consider is
$$U^E(x) = -\exp(-x) \quad (3)$$
where $E$ stands for exponential. This utility is upper bounded, and allows for unlimited assets and unlimited debts. The effective disutility of debt is exponentially growing, whereas the benefits of ever increasing assets becomes marginal. It is a concave utility function and it has one analytic property that has a simplifying effect:
$$-\exp(W - x) = -\exp(W)\exp(-x) \quad (4)$$
which says that decision regarding a change in wealth $x$ are independent of current wealth $W$.

For such a utility function, decisions do not depend on the wealth or budget of that individual. As a result in markets where all agents have exponentially decaying negative utility, the wealth of the agents is irrelevant and can be removed from the equation. This utility function is commonly just called an *exponential utility*.

### 4.2 Market Structure

The observant reader will notice the cost of goods has not yet been mentioned. This is because the cost of goods is a function of the agents' trading preferences, and the process of trade, and so it is dependent on the market structure. It is perfectly possible to have market structures whereby the cost of goods can be different for different agents. However we will make the assumption that we have a market that allows all traders to trade a given good at any time at a given cost. Hence there is associated with each good $k$ a cost $c_k$ which is the price the good is currently trading at.

We will also make a no-arbitrage assumption regarding the market and the agents in the market. That is, we will assume that it is not possible for any agent to make profitable risk free trades in a set of assets. For example if some set of goods formed a jointly certain bet, but the total price for those goods was less than one, then an agent could buy one of each of those goods and guarantee a net positive return when the bets are finalized. This is an arbitrage opportunity. If such opportunities ever arise, traders would immediately trade on those opportunities so that they quickly disappear. Any individual or group who makes himself or herself open to arbitrage trading will quickly lose money, and hence will adjust his or her position.

One of the common features of prediction markets is the association of the cost of goods in a working market with the probabilities of the outcomes associated with those goods. There are a number of good theoretical reasons why this association is valid, and it is related to betting interpretations of Bayesian inference (see e.g. [4]). For space reasons we are not able to elaborate this here, except to note that the no-arbitrage assumption ensures that for $k = 1, 2, \ldots N_G$ enumerating a set of goods associated with mutually exclusive jointly certain events, we have

$$\sum_{k=1}^{N_G} c_k = 1 \qquad (5)$$

matching the sum-to-1 assumption for probabilities. More generally the no-arbitrage assumption ensures that, if the market goods are bets on a collection of items that form a $\sigma$-field, then the costs are a probability measure on that $\sigma$-field. It is this association of price with probability that makes prediction markets a useful tool for machine learning.

We will consider equilibrium markets in this paper. Here, market equilibrium is defined by a price and an allocation such that no trader has any incentive to trade and there is no excess demand of any good. The problem of market equilibria was first formulated in Economics by Walras [23] in 1874. The existence of such a market equilibrium was established by Arrow and Debreu [1] using analysis of the fixed-point of the system.

## 5 General Formulation

Let $W_i$ denote the current wealth of agent $i$. Let the cost of goods be denoted by the cost vector $\mathbf{c} = (c_1, c_2, \ldots, c_{N_G})^T$. Then the rational agent will choose a utility maximizing position $\mathbf{s}_i = (s_{i1}, s_{i2}, \ldots, s_{ik})^T$ in each of the goods he or she has an opinion about (that is those in $S_i = \{k | m_k \in \mathfrak{F}_i\}$). However because the outcome is uncertain the actual utility of holding the goods is a weighted sum of the utility associated with each possible outcome, weighted by the agent's belief about the probability of that outcome. This is written as

$$\begin{aligned}\mathbf{s}_i^* &= \mathbf{s}_i(W_i, \mathbf{c}) \\ &= \arg\max_{\mathbf{s}_i} \sum_{j \in \Omega_i} P_i(j) U_i(W_i - \mathbf{s}_i^T \mathbf{c} + \sum_k s_{ik} r(k,j))\end{aligned} \qquad (6)$$

subject to $s_{ik} = 0$ if $k \notin S_i$. Here, $r(k, j)$ is the return of a bet on good $k$ in case of outcome $j$ and is 1 if $j \in m_k$ and zero otherwise. $\mathbf{s}_i(W_i, \mathbf{c})$ is the *buying function* for agent $i$, and states how the agent would choose to act (given no other constraints) in a market with costs $\mathbf{c}$. There is in general no guarantee that the maximum of this utility has a unique argument, and hence in general there will not be a unique buying function; additional risk free trades may be possible resulting in different purchase quantities while maintaining the same utility.

In an equilibrium market (if it exists), the agents are able to jointly act optimally given the costs $\mathbf{c}$. Hence the market constraints

$$\sum_{i=1}^{N_A} \mathbf{s}_i(W_i, \mathbf{c}) = 0 \qquad (7)$$

are satisfied for some buying functions, and can be solved to get the equilibrium costs. We will illustrate that for certain utilities these equilibrium conditions mirror known model combination procedures in machine learning.

However in a non-equilibrium situation, market or auction dynamics can also be defined. One possible dynamic is that each good comes up for auction at a time. Then all interested agents bid for those goods by giving their buying functions for that good given their holding of other goods. Costs are decided that best satisfy those bidding functions. The various bids are satisfied, and we move on to the next good etc. We will illustrate that this market dynamics has much in common with message passing schemes in probabilistic inference.

### 5.1 Discrete-state Markets

Suppose we have a prediction market, consisting of the purchase of bets that pay out 1 Grubnick on the future occurrence of one of $N_G$ mutually exclusive, jointly certain outcomes. In this case the market goods are just bets on the individual events. Note that a multivariate discrete distribution can be represented this way by enumerating all the possible joint states.

Let $W_i$ denote the current wealth of agent $i$. Let the cost of goods be denoted by the cost vector $\mathbf{c} = (c_1, c_2, \ldots, c_{N_G})^T$. Then the rational agent will choose a utility maximizing position $\mathbf{s}_i = (s_{i1}, s_{i2}, \ldots, s_{ik})^T$ in each of the goods. This is written as

$$\mathbf{s}_i^* = \mathbf{s}_i(W_i, \mathbf{c})$$
$$= \arg\max_{\mathbf{s}_i} \sum_k P_i(k) U_i(W_i - \mathbf{s}_i\mathbf{c} + s_{ik}) \quad (8)$$

where $P_i(k)$ denotes the belief of agent $i$ about the probability of the event $k$ occurring (or more accurately the event associated with good $k$ occurring). We collect these into a vector $\mathbf{p}_i = (P_i(1), P_i(2), \ldots, P_i(K))^T$.

Every agent has an opportunity for financially-neutral risk-free trades, due to the arbitrage-free assumption, by buying (or selling) one unit of every stock. The utility associate with having holding $\mathbf{s}$ and $\mathbf{s} + \alpha\mathbf{1}$ is identical: there are various equivalent positions that are produced by risk free purchases or sales. If we also introduce an additional *neutral agent* that only makes these risk free trades, buying/selling one of each item (and hence never has any difference in his/her return from the zero position), then we only need to specify the position of each agent up to these utility-equivalent classes. We introduce a *standardization constraint* to ensures each equivalence set is now represented by a single position $\mathbf{s}_i$ that satisfies the constraint. Note this does not mean that the agents have to obey the standardization constraint: that is irrelevant. It just means that any solutions we obtain that do obey the standardization constraint will be a suitable representation for all the other equivalent positions.

In the analysis we will use the most convenient constraint for any given problem. If the constraint holds for $\mathbf{s} = 0$ for all agents, then the market constraint (7) will also hold at equilibrium. One useful constraint is to set $\mathbf{s}_i^T\mathbf{c}$ to zero for each agent (each agent could buy/sell one of each stock with no change in utility until this constraint were satisfied). Another possible constraint is that we choose to set the stock holding of all agents of stock $k = N_G$ to zero. Alternatively one could require that the minimum stock holding for an agent in at least one stock was zero.

### 5.1.1 Case 1: Linear debt-free utility gives weighted median model combination

In the case of a linear debt-free utility, we use the standardization constraint $\min_k s_{ik} = 0$, which is useful in that it ensures that the active agents only make long positions (the remaining risk free short position is held by the neutral agent), and so we can write (8) as

$$\mathbf{s}_i^* = \arg\max_{\mathbf{s}_i} \mathbf{s}_i^T(\mathbf{p} - \mathbf{c})$$
$$\text{s.t. } \min_k s_{ik} = 0 \text{ and } \mathbf{s}_i^T\mathbf{c} - s_{ik} < W_i \,\forall k. \quad (9)$$

where the conditions are those imposed by the standardization constraint and the debt free constraint. Due to the standardization constraint, the last condition $\mathbf{s}_i^T\mathbf{c} - s_{ik} < W_i \,\forall k$ is satisfied if and only if $\mathbf{s}_i^T\mathbf{c} < W_i$, which simply states that the maximum stake is the whole wealth. This leads to

$$\mathbf{s}_i^* = \arg\max_{\mathbf{s}_i} \mathbf{s}_i^T(\mathbf{p} - \mathbf{c})$$
$$\text{s.t. } \min_k s_{ik} = 0 \text{ and } \mathbf{s}_i^T\mathbf{c} < W_i. \quad (10)$$

This is optimized by staking the whole wealth $W_i$ on the good $k$ which maximizes

$$\frac{P_i(k) - c_k}{c_k}. \quad (11)$$

In the binary case with many players of equivalent wealth, the equilibrium for the single cost $c$ will be the median of the agents' $P_i(1)$ values as that will balance the total long and short positions in the one good; with varying wealth, each agent's $P_i(1)$ value will be weighted by its wealth before computation of the median. Note that the fact that the linear debt free utility is not strictly convex means that there is not necessarily a unique equilibrium, which is clear from this solution as the median is not uniquely defined for even numbers of equally wealthy agents.

In multi-class settings this utility results in markets that choose costs to balance agents' purchases across all the stocks, though the exact formalism is not as simple as in the binary setting as it is dependent on the number of agents involved.

### 5.1.2 Case 2: Logarithmic utility gives weighted mean model combination

With logarithmic utility,

$$U_i^L(W_i, \mathbf{c}, \mathbf{s}_i) = \sum_k P_i(k) \log(W_i - \mathbf{s}_i^T\mathbf{c} + s_{ik}), \quad (12)$$

the market constraints (7) can also be solved. We use the standardization constraint $\mathbf{s}_i^T\mathbf{c} = 0$ to build a Lagrangian

$$L_i = \sum_k P_i(k) \log(W_i + s_{ik}) + \lambda_i \mathbf{s}_i^T\mathbf{c} \quad (13)$$

with Lagrange multiplier $\lambda_i$. By equating the derivatives of this Lagrangian to zero we get

$$\frac{\partial L_i}{\partial s_{ik}} = \frac{P_i(k)}{W_i + s_{ik}} + \lambda_i c_k = 0 \quad (14)$$

which we solve to get the buying function

$$s_{ik}^* = \frac{W_i(P_i(k) - c_k)}{c_k}. \quad (15)$$

Solving for the market constraint (7) gives the equilibrium cost

$$c_k = \frac{\sum_i W_i P_i(k)}{\sum_i W_i} \quad (16)$$

which sets the costs to be the wealth weighted mean of the agents beliefs. Note that this is a linear aggregation of classifiers, akin to methods used in boosting algorithms [9, 20] and model averaging approaches. If the wealth has been

Machine Learning Marketsachieved through past performance, then the classifiers are effectively weighted by their performance in previous circumstances, and so this also relates to a mixture of experts approach.

Note that in [15], the formula (16) was obtain via presuming constant 'betting functions' (the proportion of wealth bet as a function of cost). In reality constant betting functions are unrealistic in utility terms as they would imply always betting the same amount on the same goods irrespective of price. Here we show that a constant betting function is not necessary for a weighted mixture pricing scenario. Instead we have derived a utility consistent buying function that has the same pricing properties.

### 5.1.3 Exponential decaying negative utility gives product model combination

We can also consider the multi-class market with the exponential decaying negative utility $U^E$ The utility for agent $i$ is written as

$$U_i^E(W_i, \mathbf{c}, \mathbf{s}_i) = -\sum_k P_i(k) \exp(-W_i + \mathbf{s}_i^T \mathbf{c} - s_{ik}) \quad (17)$$

Once again, we use the standardization constraint $\mathbf{s}_i^T \mathbf{c} = 0$ to build a Lagrangian for this of the form

$$L_i = -\sum_k P_i(k) \exp(-W_i - s_{ik}) - \lambda_i \exp(-W_i) \mathbf{s}_i^T \mathbf{c} \quad (18)$$

with Lagrange multiplier expressed as $-\lambda_i \exp(-W_i)$ for convenience. By equating the derivatives of this Lagrangian to zero we get

$$\frac{\partial L_i}{\partial s_{ik}} = \exp(-W_i) P_i(k) \exp(-s_{ik}) - \lambda_i \exp(-W_i) c_k = 0 \quad (19)$$

which we solve to get the buying function

$$s_{ik}^* = \log P_i(k) - \log c_k - \log \lambda_i \quad (20)$$

where $\lambda_i$ is set to ensure $\mathbf{s}_i^T \mathbf{c} = 0$. We can now solve for (7) giving

$$c_k \propto \prod_{i=1}^{N_A} P_i(k)^{1/N_A} \quad (21)$$

which sets the costs to be the geometric mean of the agents beliefs, and is a product model with the potentials for each product scaled according to the number of agents. Hence the exponential decaying negative utility implements a product combination. By writing $\Phi_i(k) = (1/N_A) \log P_i(k)$ we have

$$c_k = \frac{1}{Z} \exp\left(\sum_i \Phi_i(k)\right) \quad (22)$$

where $Z$ is a normalization constant. We can see that each agent implements a separate contributing potential to the overall market distribution.

### 5.2 Interim Summary

We have shown how, in a market for a single multiclass outcome, many of the standard aggregation methods used for constructing componential machine learning models are reproducible as a result of different agent utility functions. Weighted median of experts, weighted mixture of experts, and product of expert models can all occur, simply by changing the utility functions involved. Furthermore, as each agent acts independently, the market mechanism provides a well defined approach for mixing agents with different utility functions together. As a result we can obtain intermediates between product distributions and mixture distributions.

In the rest of this paper we generalize the methods for multivariate settings: here there are $J$ different variables, and the goods are bets on the joint state of all these variables. Hence $N_G$ is exponential in $J$. First we consider the case of local agents, and show how this relates to factor graphs, and using a different form, for methods of combining multiple marginal beliefs. Finally we show that, when only a limited number of goods are available, market dynamics can implement message passing mechanisms to obtain the equilibrium.

## 6 Niche Agents

In economic markets, agents do not usually try to comprehend the complete joint system. Rather, individual agents establish *niches* that they attempt to exploit. In general, given some market pricing, an agent may believe that the real value differs, in some limited way, from the overall market price decided collectively by all the agents, and each agent, learns and represents his or her beliefs relative to the market, and enters the market to exploit (in his or her opinion) that difference.

There may also still be agents with direct opinion (not expressed relative to the market price) in that marketplace. We consider, for purposes of illustration, a market consisting of a single agent with a direct opinion, and a number of agents with opinions relative to the market price. The belief of each of these agents can now be represented as probability distribution that is a factorial deviation from that market price:

$$P_i(k) = \frac{1}{Z} F_i(k) c_k. \quad (23)$$

In a multivariate setting those deviations will generally occur in only a few random variables that the agents are knowledgeable about: the relative beliefs of an agent due to variations in other variables will match the distribution established by the market as a whole. For example suppose $\mathbf{y} = (y_1, y_2, \ldots, y_J)^T$ denotes the final outcome of a multivariate occurrence, where each element $y_j$ is, for the sake of notational simplicity, assumed to be binary. There are then $N_G = 2^J$ possible goods, each a bet on some out-come $\mathbf{y}$.

If agent $i$ only had deviant opinions from the general consensus regarding variables in set $S_i$, we would write $F_i(\mathbf{y}) = F_i(\mathbf{y}^{S_i})$, where we use the superscript notation $\mathbf{y}^S$ to denote the vector derived from restricting the vector $\mathbf{y}$ to just the elements with index in $S$. The set $S_i$ would be called a *clique*. Colloquially speaking agent $i$ is happy to agree with the consensus opinion regarding the variables (s)he has no knowledge about. Let $k = 1, 2, \ldots, N_G$ enumerate all the different $\mathbf{y}$: $\mathbf{y}_1, \mathbf{y}_2, \ldots, \mathbf{y}_{N_G}$. Then we use $F_i(k)$ to represent $F(\mathbf{y}_k)$ etc.

Once again, let $\mathbf{c}$ represent the cost vector (now of length $N_G = 2^J$, one term for each good). We can write out the utility for such a set of agents as

$$U_i(W_i, \mathbf{c}, \mathbf{s}_i) = \sum_{k=1}^{N_G} F_i(k) c_k U_i(W_i - \mathbf{s}_i^T \mathbf{c} + s_{ik}) \quad (24)$$

(the utility only needs to be defined up to a constant for optimization purposes). The utility for the single agent with a direct opinion can be written

$$U_0(W_i, \mathbf{c}, \mathbf{s}_i) = \sum_{k=1}^{N_G} P_0(k) U_0(W_0 - \mathbf{s}_0^T \mathbf{c} + s_{0k}) \quad (25)$$

In this case an exponential decay negative utility results in buying functions

$$s_{ik}^* = \log F_i(k) - \log \lambda_i \quad (26)$$

with Lagrange multiplier $\lambda_i$ and

$$s_{0k}^* = \log P_0(k) - \log c_k - \log \lambda_0 \quad (27)$$

The market constraint then gives an equilibrium price equation of

$$c_k \propto P_0(k) \prod_i F_i(k) \quad (28)$$

which is a product of local clique factors, along with a global factor, which represents some base distribution and could be uniform. Hence the use of agents that declare their beliefs relative to the market produces models of the form of a various local clique potentials.

The equilibrium pricing for this market represents a joint probability distribution for a the standard factor model, with each agent representing a factor over a clique of the variables. We have shown that for a certain market structure and certain utility representation, the market precisely implements a very common form of probabilistic graphical model. Any factor graph can be represented as a market of this form, and the equilibrium pricing of the market represents the probabilities associated with that factor graph.

### 6.1 Marginal Agents

The niche agents described above are interesting in terms of the model they implement. However, the majority of agent models will not result in equilibrium costs that can be simply expressed. Nevertheless, such agents could still be very valuable, as market dynamics will establish equilibria that are valid probability distributions and may satisfy desirable criteria.

One of the problems with the niche agents is that they rely on the consensus opinion of the other agents, and express their beliefs as perturbations from that opinion. That can open the agent up to taking risks entirely on the basis of the opinion of others.

Another approach is that agents may wish to purchase bets that are risk free in the variables they have no opinion about. In large systems any agent may only have knowledge about a small subsection of that system e.g. a few variables. The agent does not wish to make assumptions about the other variables. Different agents may well want to purchase bets on different subsets of those variables, due to their indifference regarding the others. We start the analysis by considering goods covering all possible multivariate states, and note that in order to make a bet on a restricted number of states, an agent need only purchase multiple equal bets covering all the options of the remaining states.

Let $\mathbf{y} = (y_1, y_2, \ldots, y_J)$ denote the final outcome of a multivariate occurrence, where each $y_j$ is, for the sake of notational simplicity, assumed to be binary. Let $S_J$ denote the collection $\{1, 2, \ldots, J\}$ of all the variable indices.

The market goods consist of bets on a payout for each $\mathbf{y}$, and hence we use the $\mathbf{y}$ to label the goods, and write $c(\mathbf{y})$ for the cost of a bet on outcome $\mathbf{y}$, and $s_i(\mathbf{y})$ for the amount of good $\mathbf{y}$ agent $i$ has.

Each agent also has a belief, but now the beliefs can be restricted to a subset $S_i$ of the variables. Once again we call the sets $S_i$ *cliques*. We will use the shorthand $\mathbf{y}_i$ to denote $\mathbf{y}^{S_i}$ where that does not cause confusion.

The rational agent will choose a utility maximizing position $s_i^*(\mathbf{y}^{S_i})$ written as

$$s_i^*(\mathbf{y}_i) = \arg\max_{s_i(\mathbf{y}_i)} \sum_{\mathbf{y}_i} P_i(\mathbf{y}_i)$$
$$U_i(W_i - \sum_{\mathbf{y}_i'} s_i(\mathbf{y}_i') c(\mathbf{y}_i') + s(\mathbf{y}_i)) \quad (29)$$

where $c(\mathbf{y}_i) = \sum_{\mathbf{y}'|(\mathbf{y}')^{S_i} = \mathbf{y}_i} c(\mathbf{y})$ is the sum of the costs of all the goods needed to produce a bet on the marginal outcome $\mathbf{y}_i$. Once again we will consider an exponential decaying negative utility.

We introduce the standardization constraint $\sum_{\mathbf{y}_i} s_i(\mathbf{y}_i) c(\mathbf{y}_i) = 0$ and optimise the agents utility with respect to this constraint to get the agent's buying function. Equating the derivative of the Lagrangian with respect to $s_i(\mathbf{y}_i)$ to zero, we get the buying function

$$s(\mathbf{y}_i) = \log P_i(\mathbf{y}_i) - \log c(\mathbf{y}_i) - \lambda_i \quad (30)$$

where a purchase of goods $\mathbf{y}_i$ consists of an equal purchase of all goods $\mathbf{y}$ consistent with $\mathbf{y}_i$ on set $S_i$. We have no simple representation for the costs of all the goods in an equilibrium market of agents with these buying functions (the number of goods is now exponential in the number of variables, and the buying functions depend on the costs for many different goods). However the market will still implement this distribution as an equilibrium of the dynamical system that defines a particular choice of market dynamics. The market will then provide a mechanism for combining a number of *marginal beliefs* about a system.

## 6.2 Message passing

Markets consisting of an exponential number of goods are practically infeasible. It will become impossible to keep track of or even represent the price of such a large number of goods. As a result the market is likely to consist only of a reduced set of the possible goods. Just as it is likely that agents will only have opinions on a small set of goods, so the market as a whole will only involve trades on a smaller set of goods than all those that are possible.

Again let $\mathbf{y} = (y_1, y_2, \ldots, y_J)$ denote the final outcome of a multivariate occurrence, where each $y_j$ is, for the sake of notational simplicity, assumed to be binary. Let $S_J$ denote the collection $\{1, 2, \ldots, J\}$ of all the variable indices. The market goods are bets on the outcome $y_j = 1$ for each $j$. The total number of goods is $N_G = J$, and so $S_J = S_{N_G}$. Each good is indexed by some $k$ chosen from the set $S_{N_G}$.

An agent will, once again, have the probabilistic belief $P_i(\mathbf{y})$. Given some market cost $c(\mathbf{y})$, and an exponential decaying negative utility, the agent will have an expected utility of

$$U_i(W_i, \mathbf{c}, \mathbf{s}_i) = -\sum_{\mathbf{y}} P_i(\mathbf{y}) \exp(-W_i + \mathbf{s}_i^T(\mathbf{c} - \mathbf{y})). \tag{31}$$

Suppose agent $i$ has been communicated all the costs for all the goods (for a marginal/niche agent this would only need to be all the costs in the clique). Then that agent is able to optimize its position in those goods to obtain a price conditional optimal value $\mathbf{s}_i^*$. We can then communicate that position in the following way:

Consider a trade in a single good $k$. Given the agents optimized position, and given the current prices $c_k$, we define $A_{ik}(y_k)$ by

$$A_{ik}(y_k) = \sum_{\mathbf{y}^{-k}} P_i(\mathbf{y}^{-k}|y_k) \exp((\mathbf{s}_i^{*-k})^T(\mathbf{c}^{-k} - \mathbf{y}^{-k})) \tag{32}$$

where the superscript $-k$ notation denotes the vector with the $k$th term removed. Then we can write $U_i(W_i, \mathbf{c}, \mathbf{s}_i^*)$ as

$$-\sum_{y_k} A_{ik}(y_k) P_i(y_k) \exp(-W_i + s_{ik}(c_k - y_k)) \tag{33}$$

where $s_{ik}$ is the holding in stock $k$.

Taking derivatives of this expected utility with respect to $s_k$, and equating to zero, we get

$$\sum_{y_k}(c_k - y_k) \exp(s_{ik}(c_k - y_k)) A_{ik}(y_k) P_{ik}(y_k) = 0 \tag{34}$$

where $P_{ik}$ is the marginal belief about $y_k$. Hence we can write the optimized position in good $k$, $s_{ik}$ as

$$s_{ik}(c_k) = \log \frac{1 - c_k}{c_k} + \log \frac{A_{ik}(1) P_{ik}(1)}{A_{ik}(0) P_{ik}(0)} \tag{35}$$

conditioned on the knowledge of the positions in the other goods.

Given this buying function, the equilibrium constraint gives

$$c_k(y_k) \propto \prod_i A_{ik}(y_k)^{1/N} P_{ik}(y_k)^{1/N} \tag{36}$$

This means we can compute the price for a bet on variable $k$ given the price of everything else, so long as we have computed the messages $A_{ik}$ for all the agents. The new cost then gets passed to all the agents so they can update their messages $A_{ik}$ resulting in new buying functions.

Though this applies generally, it is not useful unless the $A_{ik}$ are straightforward to compute. This is only the case for niche or marginal agents. In those situations (32) involves only a sum over the local clique and hence is computable in time exponential in the clique size.

## 7 Discussion

In this paper we establish the flexibility of machine learning markets for representing, through market prices, different forms of compositional machine learning model. We show that many of the compositional structures typically used in machine learning, the localized representations, and the inferential mechanisms such as message passing schemes can be interpreted in terms of machine learning markets. Put simply, certain probabilistic machine learning models can be redefined as sets of independent agents with particular utility functions. Any choice of convergent market dynamics can then be viewed as an inference approach. In this way the propagation of cost information and purchase information can be seen as messages that are passed between independent agents, much as message passing schemes work between nodes in a graph.

The benefit of this approach is that is allows for considerably more versatile models to be set up, by using multiple agents with different utility functions. These agents can function independently and need no information about what other agents are doing save for the prices they are willing to sell market goods for. This approach has significant long term appeal: it allows for immediate integration of multiple different types of agents as well as a natural, large scale parallel process for inference.

## Acknowledgements

The author would like to thank Jono Millin and Krzysztof Geras for very helpful comments for the final version of the paper, and Jono for presenting the paper at the conference.